\documentclass[10pt,twocolumn,letterpaper]{article}

\usepackage[pagenumbers]{cvpr} % To force page numbers, e.g. for an arXiv version

\usepackage{url}
\usepackage{booktabs}       % professional-quality tables
\usepackage{amsfonts}       % blackboard math symbols
\usepackage{nicefrac}       % compact symbols for 1/2, etc.
\usepackage{microtype}      % microtypography
\usepackage[table,dvipsnames]{xcolor}         % colors
\usepackage{graphicx}
\usepackage{lipsum}

\usepackage{multirow}
\usepackage{pifont}
\usepackage{enumitem}

\definecolor{cvprblue}{rgb}{0.21,0.49,0.74}
\usepackage[pagebackref,breaklinks,colorlinks,allcolors=cvprblue]{hyperref}

\usepackage{cuted}
\usepackage{booktabs}       % professional-quality tables
\usepackage{amsfonts}       % blackboard math symbols
\usepackage{nicefrac}       % compact symbols for 1/2, etc.
\usepackage{microtype}      % microtypography
\usepackage[table,dvipsnames]{xcolor}         % colors
\usepackage{graphicx}
\usepackage{lipsum}

\usepackage{multirow}
\usepackage{pifont}

\definecolor{light_gray}{gray}{0.97}
\definecolor{light_green}{RGB}{220,248,225}
\definecolor{light_blue}{RGB}{209,231,255}
\definecolor{lightblue}{RGB}{220,235,250}
\definecolor{robot_color}{RGB}{90,111,118}
\definecolor{transfer_color}{RGB}{0,112,192}
\definecolor{human_color}{RGB}{141,105,137}
\usepackage[most,skins,theorems]{tcolorbox}
\usepackage{enumitem}

\def\paperID{xxxx} % *** Enter the Paper ID here
\def\confName{CVPR}
\def\confYear{2026}

\title{\textcolor{human_color}{H}\textcolor{transfer_color}{2}\textcolor{robot_color}{R}-Grounder: A Paired-Data-Free Paradigm for Translating Human Interaction Videos into Physically Grounded Robot Videos}

\author{Hai Ci, \; Xiaokang Liu, \; Pei Yang, \; Yiren Song, \; Mike Zheng Shou\thanks{Corresponding Author}\\
Show Lab, National University of Singapore\\
{\tt\small {\{cihai03,mike.zheng.shou\}}@gmail.com}
}
\begin{document}
\maketitle
\begin{strip}
\centering
    \includegraphics[width=\linewidth]{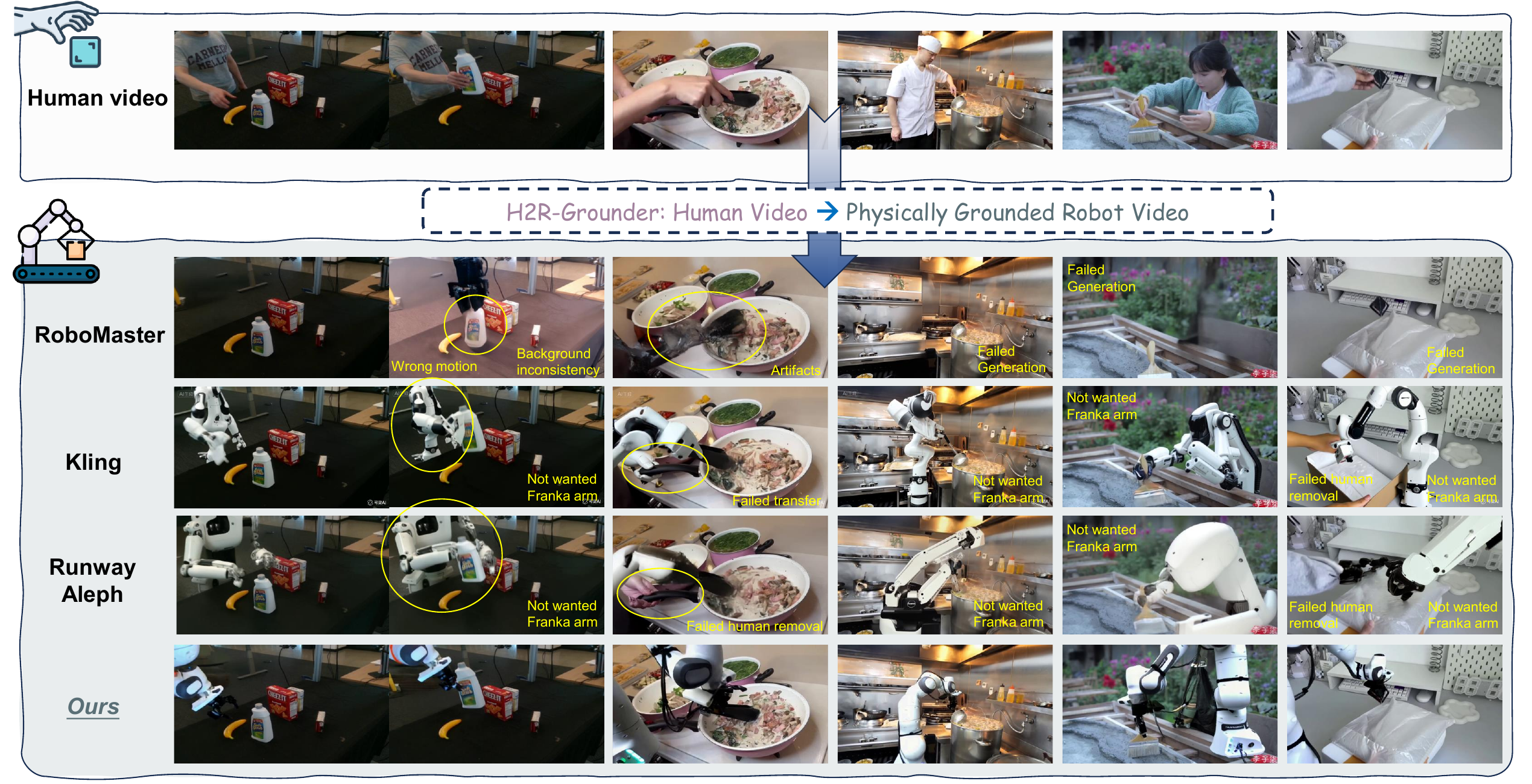}
    \captionof{figure}{\textbf{H2R-Grounder} converts human interaction videos into temporally aligned robotic manipulation videos, maintaining motion and background consistency and ensuring physically plausible robot arm structures and interactions. RoboMaster~\cite{robomaster} (animation-based) losees motion and background consistency. Kling~\cite{kling} and Runway Aleph~\cite{aleph} (editing-based) produce geometrically distorted robot arms.}
    % \vspace{-0.3em}
    \label{fig:teaser}
\end{strip}

\begin{abstract}
% \vspace{-0.2em}
Robots that learn manipulation skills from everyday human videos could acquire broad capabilities without tedious robot data collection. We propose a video-to-video translation framework that converts ordinary human–object interaction videos into motion-consistent robot manipulation videos with realistic, physically grounded interactions. Our approach does not require any paired human–robot videos for training – only a set of unpaired robot videos, making the system easy to scale. We introduce a transferable representation that bridges the embodiment gap: by inpainting the robot arm in training videos to obtain a clean background and overlaying a simple visual cue (a marker and arrow indicating the gripper’s position and orientation), we can condition a generative model to insert the robot arm back into the scene. At test time, we apply the same process to human videos (inpainting the person and overlaying human pose cues) and generate high-quality robot videos that mimic the human’s actions. 
We fine-tune a SOTA video diffusion model (Wan 2.2) in an in-context learning manner to ensure temporal coherence and leveraging of its rich prior knowledge. 
Empirical results demonstrate that our approach achieves significantly more realistic and grounded robot motions compared to baselines, pointing to a promising direction for scaling up robot learning from unlabeled human videos. Webpage: \url{https://showlab.github.io/H2R-Grounder/}
\end{abstract}    
\section{Introduction}
\label{sec:intro}
Collecting large-scale, diverse robot manipulation data remains a core challenge in robotics~\cite{rt1,pi05,droid,openxembodiment}. Recording demonstrations with physical robots is slow, costly, and constrained to lab settings~\cite{bridgev2}, leaving even the largest robot datasets far smaller and less varied than those in NLP. In contrast, human interaction videos—from casual online clips to egocentric recordings—are abundant and richly depict diverse manipulation behaviors. If robots could learn directly from these human videos, data collection would be vastly accelerated. 
Prior efforts often rely on specialized hardware~\cite{egozero} to collect paired human–robot data~\cite{hopman,egomimic} for learning, which limits scalability.
Moreover, the visual embodiment gap—human arms and hands differ significantly in appearance and motion from robot arms and grippers—makes the learning non-trivial.

\begin{figure}[t]
    \centering
    \includegraphics[width=\linewidth]{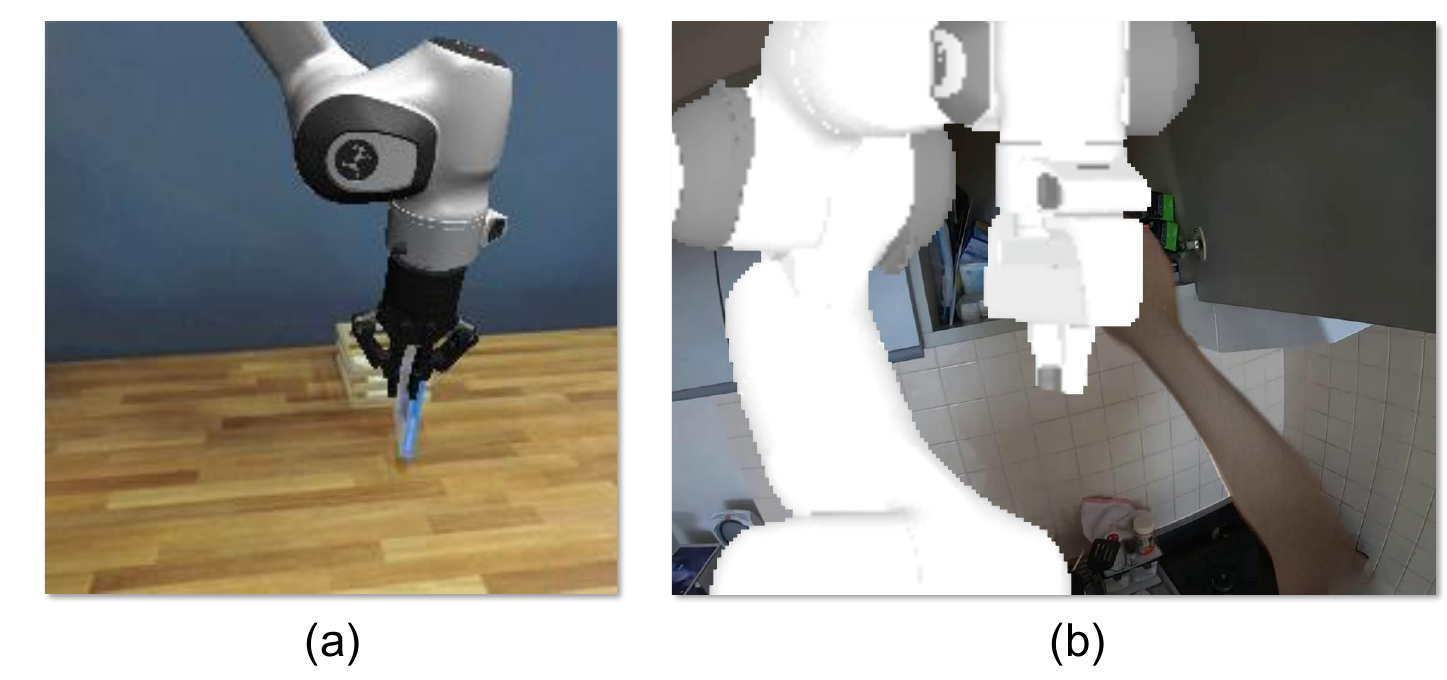}
    % \vspace{-4pt}
    \caption{Issues in prior rendering-based H2R methods. (a) shows the rendered robot arm from \textbf{Phantom}~\cite{phantom}, produced using their released code and provided calibrated camera parameters. Without accurate depth, the gripper appears to “float’’ above the book. (b) shows an overlaid robotic arm from \textbf{H2R}~\cite{h2r}, collected from their public dataset, which suffers from severe floating artifacts and camera misalignment.}
    \label{fig:floating}
\end{figure}

Recent works~\cite{phantom,masquerade,h2r} attempt to “robotize’’ human videos by rendering a robot arm into them to fill the visual gap, enabling imitation learning~\cite{masquerade} or representation learning~\cite{h2r} for policy improvement. For instance, Phantom~\cite{phantom} inpaints the human hand in video frames and overlays a rendered robot arm in its place based on the estimated hand pose. Masquerade~\cite{masquerade} and H2R~\cite{h2r} extend this idea to egocentric views. Although effective, these rendering-based methods often produce physically inconsistent visuals—robots may appear to float or misalign with objects—and require accurate camera calibration and pose estimation, which hinders generalization to in-the-wild videos. See~\cref{fig:floating}.

In this paper, we introduce \textit{H2R-Grounder}, a novel framework that marries the strengths of generative video models with a simple, transferable representation of manipulation, \textit{H2Rep}. Our key insight is to remove the need for any paired human–robot videos in training by using only unpaired robot videos and an abstract conditioning signal that is common to both human and robot domains. Concretely, we take a collection of robot manipulation videos (which may be limited in scene diversity) and algorithmically strip the robot from them: we inpaint the robot arm out of each frame, yielding a clean background video of the scene and target objects. Into this background, we overlay a minimal pose indicator – a colored dot and arrow that mark the robot gripper’s 2D location and orientation. This annotated video serves as the conditioning input. We then fine-tune a pre-trained diffusion video generator (Wan2.2~\cite{wan}) to reconstruct the original robot video given this conditioned input. Through this process, the model learns to “insert” a robot arm into a scene according to the provided pose cues, effectively learning the mapping from gripper end-effector pose sequences to realistic robot imagery. 
Crucially, the model is learning from actual robot videos, so it observes correct physics, contacts, and occlusions during training – but it never sees a human in these videos. 

At test time, we can apply the same procedure to a human demonstration video: estimate the human’s hand pose, inpaint the person from the frames, and overlay the equivalent pose indicator. This produces a transferrable representation \textit{H2Rep} of the human demonstration, to which our model can now respond by generating a robot video. The result is a robot manipulation video that follows the human’s motion in the scene, with the robot properly interacting with the objects and environment (e.g. grasping and moving objects on a table, rather than hovering unnaturally). See~\cref{fig:teaser}.

Our approach offers several advantages. It eliminates the need for paired demonstrations, leverages existing robot datasets~\cite{bridgev2,openxembodiment,droid,rt1}, and produces realistic, temporally consistent results grounded in contact physics. Moreover, our in-context fine-tuning strategy enhances temporal coherence compared to popular video-to-video pipelines such as VACE~\cite{vace}. Finally, by using minimal 2D pose indicators instead of strict 3D alignment~\cite{phantom,masquerade,h2r}, our method avoids calibration dependencies and generalizes robustly to diverse internet videos. 

To summarize, our contributions are threefold:
\begin{enumerate}
    \item \textbf{A novel human-to-robot video translation framework — H2R-Grounder}, enabling robot video generation from human demonstrations without paired data.
    \item \textbf{A simple and transferable intermediate representation — H2Rep,} unifying human and robot embodiments.
    \item \textbf{An in-context fine-tuning scheme for large diffusion video models}, improving realism and temporal consistency for physically grounded generation.
\end{enumerate}

\section{Related Work}
\noindent\textbf{Intermediate Representations for Bridging Humans and Robots.}  
Learning robot control from human videos is a long-standing challenge~\cite{dexvip,videodex,dreamtocontrol}. Due to the large visual embodiment gap between human and robot domains, most works~\cite{egomimic,hopman,manipulatorindep} rely on shared intermediate representations as surrogates for joint learning.  
EgoMimic~\cite{egomimic} masks out both human hands and robot arms to minimize appearance differences. Others~\cite{bahl2022human,manipulatorindep} inpaint manipulators and rely solely on background videos. Further studies leverage affordance maps~\cite{mendonca2023structured,bahl2023affordances,egozero}, keypoints~\cite{track2act,hat,wen2023any,li2024okami,das2021model,xiong2021learning,pointpolicy}, flow~\cite{goyal2022ifor,seita2023toolflownet}, pretrained models~\cite{hrp,hrdt}, or latent features~\cite{human2robot,immimic}.  
While these representations facilitate cross-domain learning, they seldom generate robot videos directly and thus remain limited by information loss or visual misalignment.  
Our method introduces \textit{H2Rep}, combining pose sequences and background videos to preserve both motion and scene context. Unlike prior works that only use such representations for feature alignment, we employ them to directly synthesize robot videos from human inputs, closing the visual gap.

\vspace{4pt}
\noindent\textbf{Translating Human Videos into Robot Videos.}  
Recent works attempt to directly edit human videos into robot-like ones. Phantom~\cite{phantom} overlays rendered robot arms guided by estimated hand poses, while Masquerade~\cite{masquerade} extends this to egocentric dataset epic-kitchen~\cite{epickitchen}. H2R~\cite{h2r} similarly composites simulated robot arms onto inpainted egocentric frames~\cite{ego4d}.
These rendering-based pipelines exploit large-scale human data but struggle with realism—overlaid arms ignore lighting, depth, and scene geometry, leading to implausible occlusions or contacts. 
Moreover, they require accurate camera–robot calibration and sensor parameters~\cite{masquerade, phantom}, which are unavailable for in-the-wild videos. MimicDreamer~\cite{li2025mimicdreamer,team2025gigaworld} narrows this embodiment gap via generative models, yet still conditions a generator on robot renderings, inheriting the same calibration requirement.
In contrast, we adpot a fully generative approach, synthesizing robot videos conditioned on abstract 2D pose indicators. This design inherently models occlusion and contact learned from real robot data without calibration.  
HOPMan~\cite{hopman} is related, using off-the-shelf inpainting to remove robot arms and add human hands frame-by-frame~\cite{affordancediffusion}, producing in-lab human–robot pairs. However, the reverse process—translating in-the-wild human videos into robot videos—remains infeasible due to the lack of a robot video generator. Our work fills this gap by introducing such a generator.

\vspace{4pt}
\noindent\textbf{Cross-Robot Embodiment Transfer.}  
Several studies~\cite{shadow, roviaug} investigate transferring across robots with similar morphology, benefiting from their comparable kinematics.  
In contrast, our human-to-robot setting involves third-person videos with full-body humans and robotic manipulators of vastly different structures, making embodiment transfer substantially more challenging.

\vspace{4pt}
\noindent\textbf{Generative Robot Video Prediction.}  
Robot video prediction models typically generate future frames conditioned on robot actions such as 3D end-effector poses~\cite{irasim,ivideogpt,cosmos,cosmospredict25,hma,uva,worldvla,unipi,dreamgen}.  
Our generative model instead conditions on easily obtained 2D pose sequences and background videos, enforcing both pose-consistent motion and scene coherence.  
The closest baseline, RoboMaster~\cite{robomaster}, animates robot–object interaction videos from a single image given user-defined 2D robot and object trajectories, but it requires manual annotations for object masks and trajectories. We adapt RoboMaster to our H2R setting and show that H2R-Grounder achieves superior motion–background consistency and overall realism.

\begin{figure*}[!ht]
    \centering
    \includegraphics[width=0.95\linewidth]{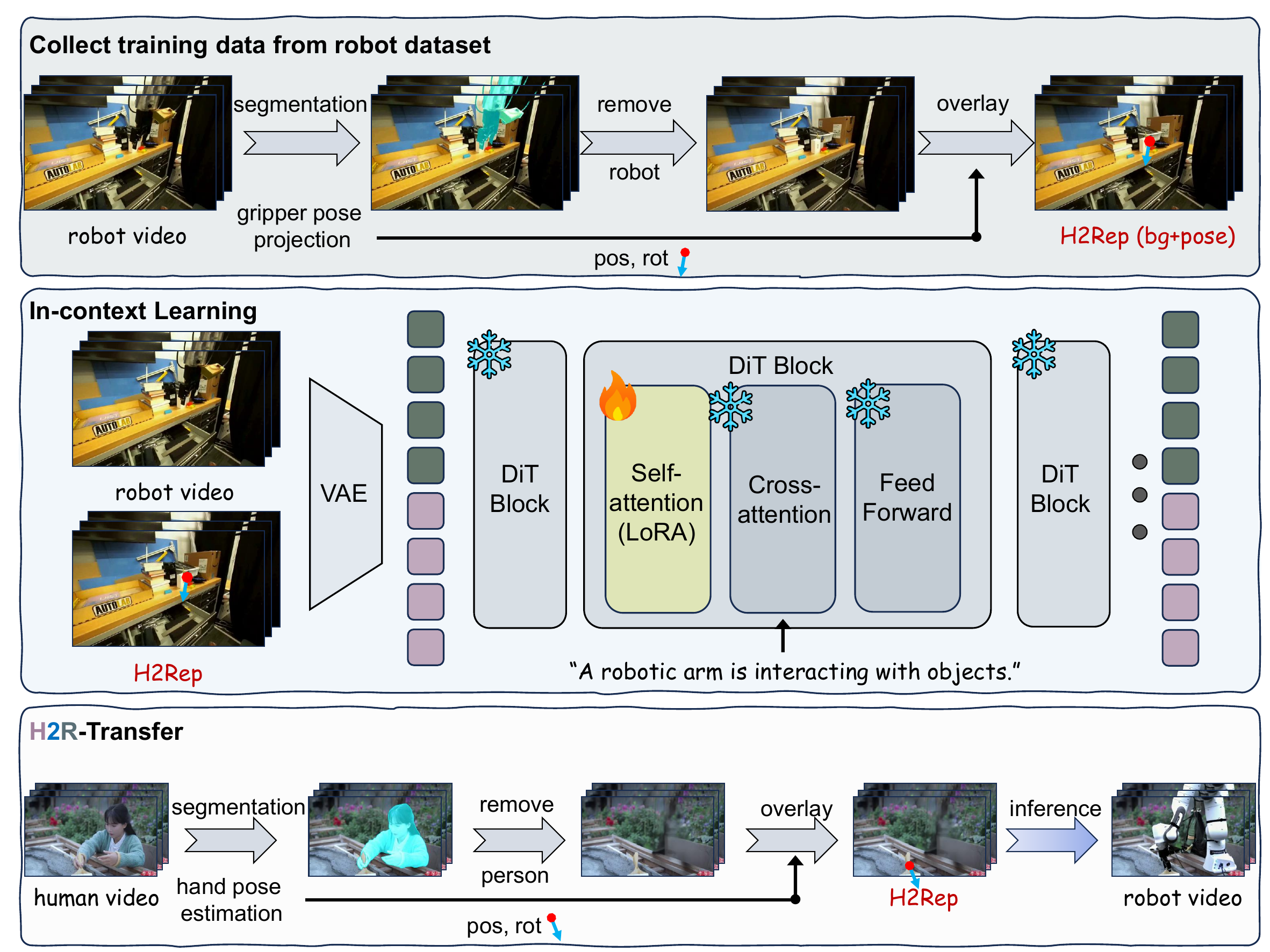}
    \caption{\textbf{Paradigm of H2R-Grounder.} The overall pipeline consists of three stages: (1) training data collection from robot video datasets, (2) in-context fine-tuning of the video generation model, and (3) transfer from in-the-wild human videos to robot manipulation videos.} 
    \label{fig:pipeline}
\end{figure*}

\section{Methodology}
\label{sec:method}
\subsection{A shared abstraction for human and robot videos}
There exist abundant human--object interaction (HOI) videos on the web and large collections of robot manipulation videos captured in labs \cite{droid,openxembodiment,bridgev2}. However, collecting \emph{frame-aligned} human--robot pairs at scale is prohibitively costly. We therefore seek a \emph{shared representation} that bridges large-scale HOI videos and robot manipulation videos without requiring paired, frame-aligned supervision.
We observe that both domains decompose naturally into: (i) a \emph{pose trajectory} of the manipulator (human hand or robot gripper) that carries action semantics, and (ii) a \emph{background video} that preserves scene layout and the physical state of manipulated objects. If we align human-hand and robot-gripper poses, then ``pose sequence + background'' becomes a common carrier of the key information in both domains. We denote this abstraction by \textbf{H2Rep}.
In the following sections, we present: (1) how to extract \textit{H2Rep}, from robot manipulation videos~(\cref{subsec:robot_h2rep}); (2) how to train an in-context video generation model conditioned on this structured representation to synthesize robot videos~(\cref{subsec:icl_train}); and (3) how to obtain \textit{H2Rep} from human–object interaction videos and leverage the video generator to generate frame-aligned robot videos~(\cref{subsec:human2robot}).
The overall three-stage pipeline is illustrated in~\cref{fig:pipeline}.

\vspace{0.25em}
\noindent\textbf{Notation.}
Let $\mathbf{V}_r$ and $\mathbf{V}_h$ be a robot video and a human video, respectively. $\mathbf{H}_r$ and $\mathbf{H}_h$ are \textit{H2Rep} extracted from robot video and human video, respectively.  We use
$\mathcal{S}$ for text-prompted video segmentation (Grounded-SAM2~\cite{gsam}), $\mathcal{I}$ for video object removal (inpainting), $\Pi$ for 6-DoF-to-2D pose projection using calibrated cameras, $\mathcal{R}$ for rendering a pose as graphic overlays (red dot for position and blue arrow for orientation), and $\mathrm{Blend}(\mathbf{A},\mathbf{B};\alpha)=(1-\alpha)\mathbf{A}+\alpha\mathbf{B}$ for alpha blending with $\alpha\!=\!0.4$. We use a video VAE encoder/decoder $(\mathsf{Enc},\mathsf{Dec})$, and $e(\cdot)$ for a text embedding.

\subsection{Training data construction from robot videos}
\label{subsec:robot_h2rep}
\paragraph{Robot-arm segmentation.}
Given a robot video $\mathbf{V}_r$, we obtain a pixel-accurate mask sequence with a text prompt:
\begin{equation}
\mathbf{M}_r \;=\; \mathcal{S}(\mathbf{V}_r,~\text{``robotic arm''}).
\label{eq:seg_robot}
\end{equation}

\paragraph{Gripper pose projection.}
Let the end-effector (EEF) 6-DoF trajectory be $\mathbf{T}_{\!\text{EEF}}(t) = [\mathbf{p}(t),\mathbf{R}(t)]$ and camera intrinsics/extrinsics be $(\mathbf{K},\mathbf{R}_c,\mathbf{t}_c)$. We project to image space:
\begin{equation}
\mathbf{P}_r(t) \;=\; \Pi\big(\mathbf{K},\mathbf{R}_c,\mathbf{t}_c;\, \mathbf{p}(t),\mathbf{R}(t)\big),
\label{eq:proj_robot}
\end{equation}
and render a dot/arrow overlay $\mathcal{R}(\mathbf{P}_r)$ on each frame.

\begin{figure}[t]
    \centering
    \includegraphics[width=\linewidth]{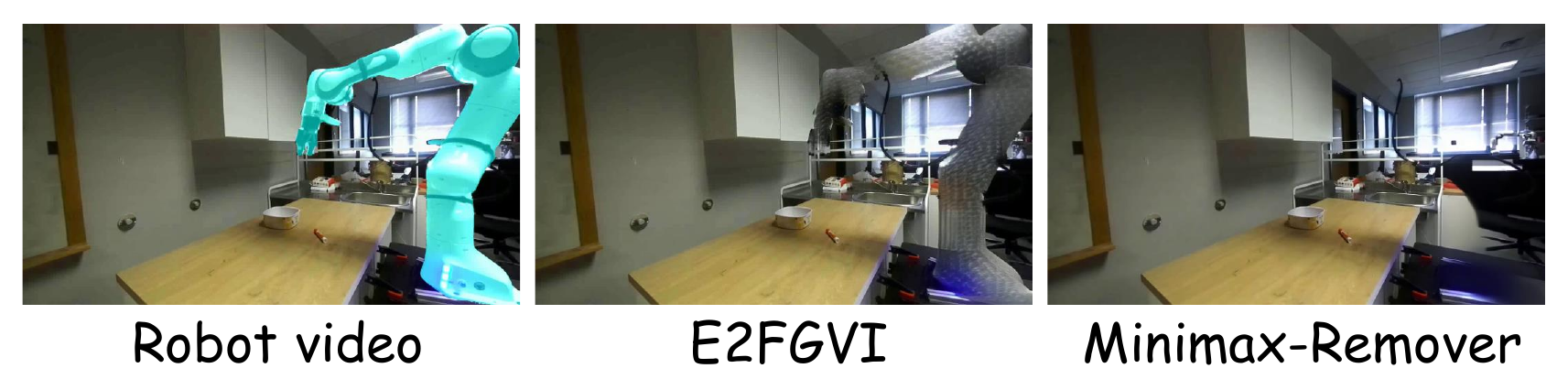}
    \caption{\textbf{Comparison of video inpainting methods} on the robot arm removal task, evaluated on a sample from the Droid~\cite{droid} dataset.}
    \label{fig:inpaint_cmp}
\end{figure}

\paragraph{Robot-arm removal (background video).}
We remove the arm with a video inpainting model:
\begin{equation}
\mathbf{V}_r^{\mathcal{I}} \;=\; \mathcal{I}\big(\mathbf{V}_r,\,\mathbf{M}_r\big).
\label{eq:inpaint_robot}
\end{equation}
Empirically, Minimax-Remover~\cite{minimaxremover} preserves background and removes the robot arm more reliably  than another popular inpainting model E2FGVI~\cite{e2fgvi}, so we adopt it in our pipeline. See~\cref{fig:inpaint_cmp}.

\paragraph{Composing robot video H2Rep.}
We form the shared representation by blending the rendered pose with the inpainted background:
\begin{equation}
\mathbf{H}_r \;=\; \mathrm{Blend}\!\left(\mathbf{V}_r^{\mathcal{I}},~\mathcal{R}(\mathbf{P}_r);\;\alpha\right),\quad \alpha=0.4.
\label{eq:h2rep_robot}
\end{equation}
This yields training pairs $\mathcal{D}_r=\big\{(\mathbf{H}_r^{(i)},\mathbf{V}_r^{(i)})\big\}_{i=1}^{N}$, where $\mathbf{H}_r$ carries gripper motion and scene evolution, and $\mathbf{V}_r$ is the physically grounded target.

\subsection{In-context learning for physically grounded robot video generation}
\label{subsec:icl_train}

We train a conditional video generator $G_{\theta}$ (Wan~2.2 backbone~\cite{wan}) to synthesize $\mathbf{V}_r$ conditioned on $\mathbf{H}_r$ (and a fixed text prompt $c_{\text{text}}$: ``A robotic arm is interacting with objects.''). Following an in-context learning design, both $\mathbf{H}_r$ and $\mathbf{V}_r$ are encoded by the same VAE and fused by self-attention; only LoRA adapters~\cite{lora} on the Q/K/V projections are trainable, while all other backbone weights remain frozen:
\begin{equation}
\mathbf{z}_H=\mathsf{Enc}(\mathbf{H}_r),\;\; \mathbf{z}_V=\mathsf{Enc}(\mathbf{V}_r),\;\;
\mathbf{c}=\big[\mathbf{z}_H;\,e(c_{\text{text}})\big].
\label{eq:vae_tokens}
\end{equation}
We adopt a flow-matching objective. Let $\mathbf{z}_0\!=\!\mathbf{z}_V$, sample $\mathbf{z}_1\!\sim\!\mathcal{N}(\mathbf{0},\mathbf{I})$, and linearly interpolate $\mathbf{z}_t\!=\!(1-t)\mathbf{z}_0+t\mathbf{z}_1$ with target velocity $\mathbf{v}_t\!=\!\frac{d\mathbf{z}_t}{dt}=\mathbf{z}_1-\mathbf{z}_0$. We train the conditional velocity field $u_\theta$:
% \begin{equation}
% \mathcal{L}_{\text{FM}}
% \,=\,
% \mathbb{E}_{t\sim\mathcal{U}(0,1)}\;
% \mathbb{E}_{(\mathbf{H}_r,\mathbf{V}_r)\sim\mathcal{D}_r}\;
% \mathbb{E}_{\mathbf{z}_1\sim\mathcal{N}}
% \Big[
% \big\| G_\theta(\mathbf{z}_t,\,t,\,\mathbf{c}) - (\mathbf{z}_1-\mathbf{z}_0) \big\|_2^2
% \Big].
% \label{eq:flow_matching}
% \end{equation}
\begin{equation}
\mathcal{L}
\,=\,
\mathbb{E}_{t\sim\mathcal{U}(0,1),\;(\mathbf{H}_r,\mathbf{V}_r)\sim\mathcal{D}_r,\;
\mathbf{z}_1\sim\mathcal{N}
}
\Big[
\big\| u_\theta(\mathbf{z}_t,\,t,\,\mathbf{c}) - \mathbf{v}_t \big\|_2^2
\Big].
\label{eq:flow_matching}
\end{equation}
% At inference, integrating $u_\theta$ from $t\!=\!1\!\rightarrow\!0$ transports a noise sample to a conditional sample; decoding gives the video:
% \begin{equation}
% \widehat{\mathbf{V}}_r \;=\; \mathsf{Dec}\!\left(\mathrm{SolveFlow}\big(u_\theta,\,\mathbf{z}_1,\,\mathbf{c}\big)\right),\quad \mathbf{z}_1\sim\mathcal{N}(\mathbf{0},\mathbf{I}).
% \label{eq:inference_dec}
% \end{equation}
At inference, robot videos $\widehat{\mathbf{V}}_r$ can be generated with the trained generator $G_{\theta}$ from robot video \textit{H2Rep} $\mathbf{H}_r$:
\begin{equation}
\widehat{\mathbf{V}}_r = G_{\theta}(\mathbf{H}_r, \mathbf{z}_1, t, \mathbf{c}_{\text{text}}).
\label{eq:robot_generation}
\end{equation}
Our H2R-Grounder ensures supervision $\mathbf{V}_r$ comes from real robot videos with genuine physical interactions, encouraging physically plausible generations.

\subsection{Human video $\rightarrow$ robot video}
\label{subsec:human2robot}

Given an arbitrary third-person HOI video $\mathbf{V}_h$, we construct its \textit{H2Rep} and feed it to the trained generator.
\paragraph{Person segmentation and hand pose.}
For any given HOI video \( V_h \), we first employ Grounded-SAM 2.1 to obtain its mask sequence \( M_h \). 
Meanwhile, we use ViT-Pose~\cite{vitpose} to estimate the human body pose and locate the hand bounding box \( B_h \), followed by HaMeR~\cite{hamer} to accurately estimate the hand pose \( P_{\text{hand}} \). 
We then take the midpoint between the index fingertip and thumb tip as the hand position, and the direction of the thumb as its orientation, forming a surrogate pose \( P_h \) that effectively represents the hand’s spatial position and direction. 
Empirically, we find that \( P_h \) aligns well to serve as a surrogate for the projected gripper pose in robot manipulation videos.
% \begin{align}
% \begin{split}
% \mathbf{M}_h \;&=\; \mathcal{S}(\mathbf{V}_h,~\text{``person''}), 
% \\
% \mathbf{B}_h \;&=\; \mathcal{D}(\mathbf{V}_h)
% \quad\text{(hand detector via ViTPose)}, \\
% \mathbf{P}_{\text{hand}} \;&=\; \mathcal{H}(\mathbf{B}_h)
% \quad\text{(hand pose via HaMeR)}, \\
% \mathbf{P}_h \;&=\; \mathcal{G}(\mathbf{P}_{\text{hand}})
% \quad\text{(surrogate 2D pose)}.
% \label{eq:hand_pipeline}
% \end{split}
% \end{align}
\begin{align}
\begin{split}
\mathbf{M}_h \;&=\; \mathcal{S}(\mathbf{V}_h,~\text{``person''}), 
\\
\mathbf{P}_h \;&=\; \mathcal{D}(\mathbf{V}_h)
\quad\text{(estimate surrogate 2D hand pose)}. \\
\label{eq:hand_pipeline}
\end{split}
\end{align}
\paragraph{Person removal (background video).} We use Minimax-Remover to remove person from the video:
\begin{equation}
\mathbf{V}_h^{\mathcal{I}} \;=\; \mathcal{I}\big(\mathbf{V}_h,\,\mathbf{M}_h\big).
\label{eq:inpaint_human}
\end{equation}
\paragraph{Composing human video H2Rep .} \textit{H2Rep} from the human video also follows the same format as from the robot video:
\begin{equation}
\mathbf{H}_h \;=\; \mathrm{Blend}\!\left(\mathbf{V}_h^{\mathcal{I}},~\mathcal{R}(\mathbf{P}_h);\;\alpha\right),\quad \alpha=0.4.
\label{eq:h2rep_human}
\end{equation}
\paragraph{H2R translation .}
We directly condition the trained robot generator $G_{\theta}$ on the human video abstract $\mathbf{H}_h$ to generate the robot video from human video:
% \begin{equation}
% \widehat{\mathbf{V}}_r^{(h)} \;=\; \mathsf{Dec}\!\left(\mathrm{SolveFlow}\big(u_\theta,\,\mathbf{z}_1,\,[\mathsf{Enc}(\mathbf{H}_h);\,e(c_{\text{text}})]\big)\right)\;\;.
% \label{eq:h2r_gen}
% \end{equation}
\begin{equation}
\widehat{\mathbf{V}}_r = G_{\theta}(\mathbf{H}_h, \mathbf{z}_1, t, \mathbf{c}_{\text{text}}).
\label{eq:h2r_translation}
\end{equation}
Because we fine-tune only lightweight LoRA adapters and keep the base generator frozen, $G_\theta$ maintains strong OOD generalization so we can apply it to in-the-wild videos.

% \subsection{A shared representation between human video and robot video}

% \cite{droid,openxembodiment,bridgev2}

% \subsection{Training data collection}
% \begin{equation}
%     M_r = \text{GSAM} \left(V_r, \text{``Robotic Arm''}\right)
% \end{equation}

% \begin{equation}
%     P_r = \text{Project2d} \left(\text{EEF pose}, \text{Extrinsics}\right)
% \end{equation}

% \begin{equation}
%     V_r^I = \text{Minimax-Remover}\left(V_r, M_r \right)
% \end{equation}

% \begin{equation}
%      H2Rep_r = \text{AlphaBlend}\left(V_r^{I}, P_r , alpha=0.4\right)
% \end{equation}

% \subsection{In-context learning for physically grounded robot video generation}

% \subsection{Human video to robot video translation}

\begin{figure*}[!ht]
    \centering
    \includegraphics[width=\linewidth]{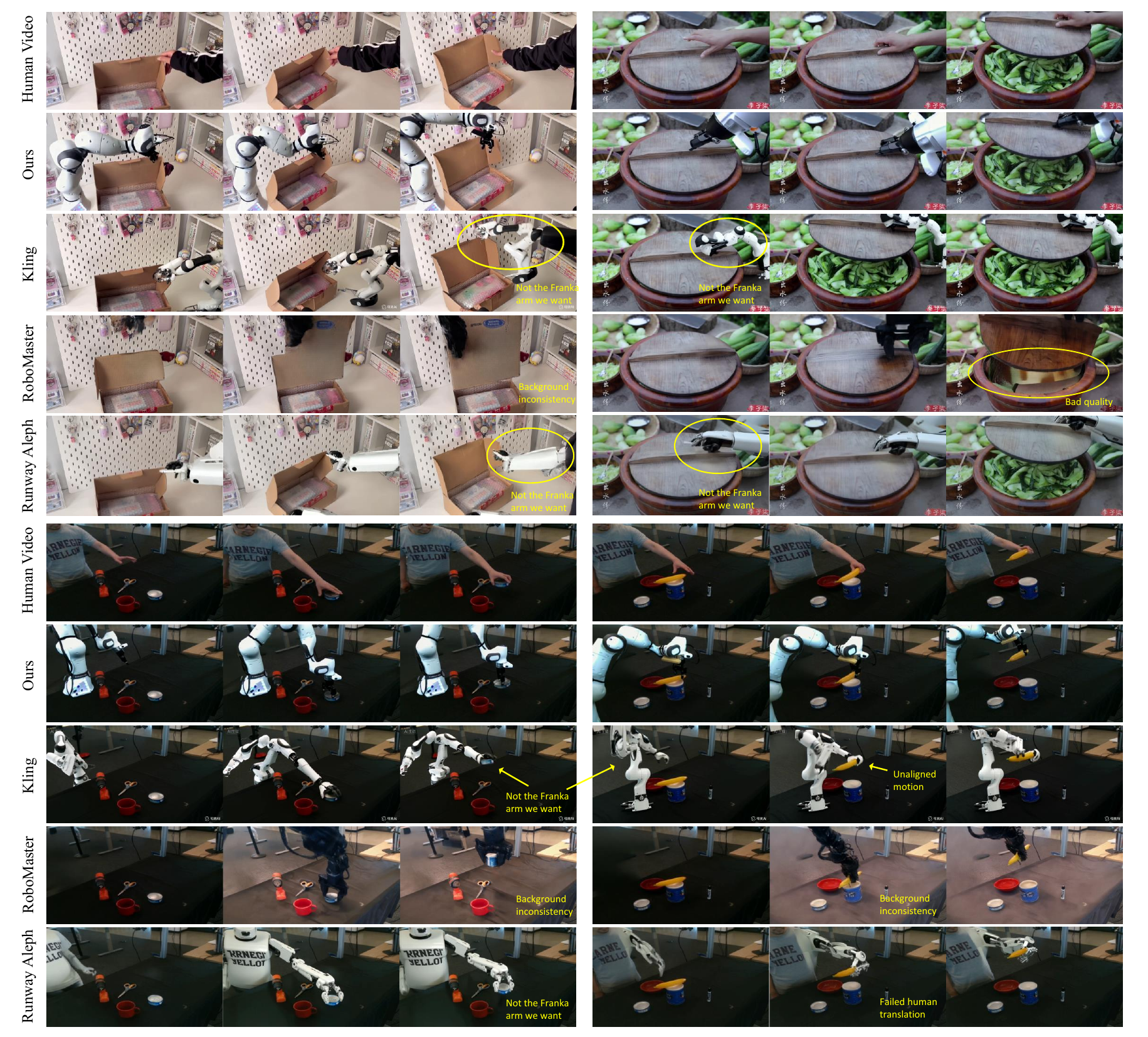}
    \caption{\textbf{OOD H2R transfer}. Top row: results on internet videos. Bottom row: results on DexYCB~\cite{dexycb} videos.} 
    \label{fig:qualitative}
\end{figure*}

\begin{figure*}[!ht]
    \centering
    \includegraphics[width=0.95\linewidth]{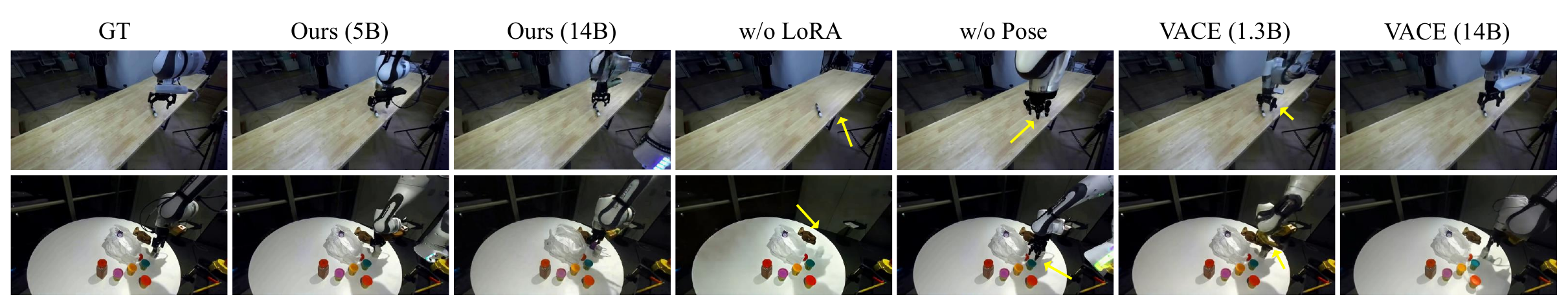}
    \caption{\textbf{Ablation on Droid~\cite{droid} data}. Examples obtained by extracting frames at the same timestep.}
    \label{fig:ablation}
\end{figure*}

\section{Experiments}
\subsection{Experimental Setup}
\vspace{4pt}
\noindent\textbf{Training and testing datasets.}  
We use the Droid dataset~\cite{droid} for training. This dataset contains approximately 76K diverse third-person Franka arm~\cite{Franka} manipulation videos. During training, we randomly sample from the whole dataset while reserving 50 for validation.  
We report SSIM, and LPIPS~\cite{lpips} to evaluate motion and background consistency as well as high-level visual feature distance between generated and ground-truth videos.  

To evaluate H2R-Grounder on out-of-distribution (OOD) human videos, we test on two types of data.  
First, we use the \textbf{DexYCB} human–object interaction dataset~\cite{dexycb}, which captures controlled lab-environment videos but exhibits clear domain shifts in both background and action distributions compared with Droid.  
It includes eight third-person camera views showing interactions between subjects and 20 distinct objects. We use the 100 videos from subject 01 under the camera 932122062010 top-down view as our test set.  
We do not use the ground-truth human masks or object poses provided by DexYCB; instead, we employ our automatic annotation pipeline described earlier to simulate real-world testing conditions.  
Since no ground-truth robot videos exist for comparison, we evaluate this set using two complementary metrics: (1) VLM-based evaluation and (2) human studies (gold standard), focusing on four aspects—motion consistency, background consistency, visual quality, and physical plausibility (robot integrity and contact realism).  In addition, we collect \textbf{internet videos} featuring more diverse backgrounds, occlusions, viewpoints, and camera motions for qualitative comparisons with baseline methods.

\vspace{4pt}
\noindent\textbf{Data preprocessing.}  
All training videos are standardized to a resolution of 1280$\times$720 and downsampled to 10\,fps.  
We trim each clip to ensure its frame count $n$ satisfies $n \bmod 4 = 1$, which is required by both Minimax-Remover and Wan for frame-aligned generation.  
During fine-tuning, we randomly sample a clip of up to 49 frames from each training video.  
Thanks to Wan’s strong pretraining, the fine-tuned generator generalizes well to human videos of different frame rates during inference.

\vspace{4pt}
\noindent\textbf{Backbone.}  
We fine-tune the Wan 2.2 TI2V-5B model~\cite{wan} as our primary video generator.  
Our H2R-Grounder establishes a novel paradigm for translating human videos into robot manipulation videos. Under this paradigm, the video generator can be replaced with other conditional video generation frameworks. We study another popular generator VACE~\cite{vace}, which adopts a ControlNet~\cite{controlnet}-based conditioning mechanism instead of in-context learning. Since VACE depends heavily on accurate textual descriptions, we additionally use Qwen2.5-VL~\cite{qwen2} to automatically generate detailed captions for all training and testing videos. We fine-tune our in-context model for 200 steps with a mini-batch size of 4, using 8 NVIDIA H200 GPUs and a gradient accumulation factor of 2. For VACE, we train for 2 full epochs on the entire dataset to ensure convergence.

\begin{table*}[!ht]
\centering
\caption{\textbf{Human preference rate on DexYCB.} Users are asked to rank the three generated videos, and our model is most frequently selected as the top choice for all aspects.}
\begin{tabular}{lcccc}
\toprule
 & Motion Consistency & Background Consistency & Visual Quality & Physical Plausibility \\
\midrule
RoboMaster~\cite{robomaster} & 2.3\% & 2.3\% & 2.3\% & 18.2\% \\
Runway Aleph~\cite{aleph} & 22.7\% & 15.9\% & 9.1\% & 6.8\%  \\ 
Kling~\cite{kling} & 9.1\% & 34.1\% & 40.9\% & 9.1\% \\
\rowcolor{cvprblue!10} Ours & \textbf{54.5\%} & \textbf{56.8\%} & \textbf{61.4\%} & \textbf{63.6\%} \\
\bottomrule
\end{tabular}
\label{tab:human_eval}
\end{table*}
\begin{table*}[!ht] 
\centering 
\caption{\textbf{VLM scoring on DexYCB.} We prompt Gemini \cite{gemini} to rate the generated videos across four aspects. Our model outperforms the baselines on most metrics, with a slight drop in visual quality compared to Kling \cite{kling}.} 
\begin{tabular}{lcccc} 
\toprule 
& Motion Consistency & Background Consistency & Visual Quality & Physical Plausibility\\ 
\midrule 
RoboMaster~\cite{robomaster} & 2.6 & 4.5 & 3.5 & 2.8 \\ 
Runway Aleph~\cite{aleph} & \textbf{3.7} & 4.5 & 3.6 & 3.9 \\ 
Kling~\cite{kling} & 3.5 & \textbf{4.9} & \textbf{4.1} & 3.6 \\ 
\rowcolor{cvprblue!10} Ours & \textbf{3.7} & \textbf{4.9} & 4.0 & \textbf{4.4} \\ 
\bottomrule
\end{tabular} 
\label{tab:vlm_eval} 
\end{table*}

\subsection{Comparison with Baselines}
\subsubsection{Rendering-Based Methods}
Rendering-based approaches such as Phantom~\cite{phantom} and Masquerade~\cite{masquerade} require precise hand–robot calibration to compute the transformation between the camera and robot frames, as well as accurate camera intrinsics and vertical field-of-view parameters for physically correct rendering of robot arms.  
Such parameters are unavailable for our in-the-wild human videos, making direct comparison infeasible. Therefore, these methods are excluded from evaluation.

\subsubsection{Animation-Based Methods}
We adapt the recently proposed robot I2V method RoboMaster~\cite{robomaster} to the human-to-robot (H2R) translation setting. The original system animates robot–object interaction videos from a static image given user-defined robot and object trajectories. To enable comparison under our setup, we construct the required inputs through a semi-manual process:  
(1) The first frame of the human–object video is inpainted to remove the human, serving as the reference frame;  
(2) hand pose trajectories are extracted following our H2R-Grounder pipeline and used as surrogates for robot trajectories;  
(3) the interacted object is manually selected and segmented using SAM 2.1~\cite{sam2} to obtain its mask;  
(4) its motion trajectory is tracked by CoTracker3~\cite{cotracker3};  
(5) the trajectory is manually divided into pre-interaction, interaction, and post-interaction phases; and  
(6) a textual caption describing the robot motion is written.  
This process allows RoboMaster to generate robot–object interaction animations, albeit with heavy manual preparation.

\subsubsection{Commercial Video-Editing Methods}
Commercial video-editing systems such as Kling~\cite{kling} and Runway Aleph~\cite{aleph} can replace the subject of a video while roughly maintaining temporal coherence and background appearance.  
We upload an image of a Franka robotic arm and prompt Kling to replace the human in each input video with the robot arm. For Aleph, we similarly prompt it to replace the human in the video with a Franka robotic arm. This serves as a practical baseline representing appearance-level subject replacement rather than true generative translation.

\subsubsection{Quantitative Results on DexYCB}
\cref{tab:human_eval} and~\cref{tab:vlm_eval} summarize the results on the DexYCB test set. We evaluate H2R-Grounder, Kling, and RoboMaster through both human studies and VLM-based scoring.

\noindent\textbf{Human study.}
We conduct a user study with 22 participants, all holding computer-science backgrounds (bachelor’s, master’s, or PhD). Each participant ranks the outputs from the three methods in terms of motion consistency, background consistency, visual quality, and physical plausibility (measured by structure integrity and contact realism).
We report the first-rank rate—the percentage of participants who selected a method as best for each aspect. Ties are allowed in the ranking, so the total percentages may not sum to 100\%.

As shown in~\cref{tab:human_eval}, \textit{H2R-Grounder} achieves the highest first-rank preference across all four evaluation aspects. It is most favored in visual quality (61.4\%) and physical plausibility (63.6\%), indicating that our generated videos are both visually convincing and physically coherent, with accurate object contacts. The high preference in motion consistency (54.5\%) and background consistency (56.8\%) further demonstrates that our model produces temporally stable motions while preserving contextual alignment.

Kling ranks second, benefiting from its commercial editing pipeline, which yields visually appealing results (40.9\%) and stable backgrounds (34.1\%). However, it struggles with motion consistency (9.1\%) and physical plausibility (9.1\%), where the synthesized arms often lose structure or exhibit implausible interactions. Runway Aleph achieves moderate results, particularly in motion consistency (22.7\%), but remains less realistic overall. RoboMaster performs the weakest, with preference rates around 2–3\% across most aspects, showing that manually defined trajectories fail to capture natural motion or consistent visual quality. Overall, the human study demonstrates that H2R-Grounder achieves the best balance between motion realism, physical grounding, and visual fidelity, without relying on paired data or calibration.

\noindent\textbf{VLM evaluation.}
We further evaluate using Gemini~\cite{gemini}, a multimodal visual–language model, to rate each generated video on a 1–5 scale across the same four criteria (Table~\ref{tab:vlm_eval}). The VLM results align with human preferences: H2R-Grounder attains the highest or comparable scores in motion consistency (3.7), background consistency (4.9), and physical plausibility (4.4), confirming its robust understanding of scene dynamics and contact physics. Kling achieves slightly higher visual quality (4.1 vs.\ 4.0), likely due to its polished rendering style, but lags behind in realism-related aspects. 
RoboMaster again performs the worst, limited by its predefined, non-adaptive motion generation. Together, these results highlight that H2R-Grounder delivers the most balanced and physically grounded video generation among all baselines.

\begin{table}[h!]
\centering
\caption{\textbf{Quantitative ablation} on the Droid dataset. $\uparrow$ indicates higher is better; $\downarrow$ indicates lower is better.}
\begin{tabular}{lcc}
\hline
 & SSIM $\uparrow$ & LPIPS $\downarrow$ \\
\hline
\rowcolor{cvprblue!10} HR-Grounder 5B (ours) & \textbf{0.82} & \textbf{0.22} \\
\hline
w/o pose indicator & 0.80 & 0.23 \\
w/o LoRA & 0.80 & 0.26 \\
w/ 14B backbone & 0.79 & 0.23 \\
w/ VACE~\cite{vace} (1.3B) & 0.68 & 0.30 \\
w/ VACE~\cite{vace} (14B) & 0.71 & 0.27 \\
\hline
\end{tabular}
\label{tab:ablation}
\end{table}

\subsubsection{Qualitative Results}
\cref{fig:qualitative} presents qualitative comparisons of H2R-Grounder against existing baselines on both internet videos and DexYCB sequences. Although our video generator is fine-tuned only on the DROID indoor dataset, it generalizes well to in-the-wild videos, maintaining consistent backgrounds, accurate motion alignment, and sharp visual quality across different viewpoints. In contrast, Kling and Runway Aleph often produces structurally inconsistent robot arms that deviate from real-world kinematics, while RoboMaster significantly distorts the background and fails to follow the demonstrated motion precisely. As shown in the bottom-right example, H2R-Grounder accurately positions the gripper to grasp the banana tip, faithfully following the human hand trajectory.

\subsection{Ablation Study}
\cref{tab:ablation} and~\cref{fig:ablation} analyze the effect of key components in H2R-Grounder.  
% Among all metrics, SSIM and LPIPS align best with human perceptual judgments, reflecting overall visual realism and structural consistency.
Removing the pose indicator from \textit{H2Rep} leads to noticeable motion drift: the generated robot arm often deviates from the intended trajectory, confirming that the pose cue is essential for motion control.  
Without LoRA fine-tuning, the model tends to overfit and does not generate an robot arm.  
Replacing the in-context video generator with VACE yields lower SSIM and higher LPIPS, showing that ControlNet-based conditioning is less effective for maintaining motion–background coherence.
Scaling to a 14B backbone does not yield clear quality improvements but drastically slows inference and limits sequence length (49 → 17 frames).  
Considering both accuracy and efficiency, we adopt the 5B model with in-context learning as our final configuration.

\section{Conclusion and Limitation}
We presented H2R-Grounder, a paired-data-free framework that translates human interaction videos into physically grounded robot manipulation videos. Leveraging the unified representation \textit{H2Rep}, our approach effectively bridges the visual embodiment gap and generates motion-consistent, realistic robot videos without calibration or paired supervision. 

\noindent
\textbf{Limitation.} Currently, the framework supports only single-hand to single-arm translation. Extending it to bimanual scenarios is feasible with appropriate dual-arm robot data and will be explored in future work. Moreover, as training is conducted solely on datasets featuring the Franka robot arm, H2R-Grounder currently produces only Franka-style outputs. Adapting to other robot embodiments would require fine-tuning or training lightweight LoRA adapters for each robot type.
{
    \small
    \bibliographystyle{ieeenat_fullname}
    \bibliography{main}

@String(IJCV = {Int. J. Comput. Vis.})

@String(CVPR= {IEEE Conf. Comput. Vis. Pattern Recog.})

@String(ICLR = {Int. Conf. Learn. Represent.})

@String(IJCV  = {IJCV})

@String(CVPR  = {CVPR})

@String(ICLR  = {ICLR})

@misc{vace,
      title={VACE: All-in-One Video Creation and Editing}, 
      author={Zeyinzi Jiang and Zhen Han and Chaojie Mao and Jingfeng Zhang and Yulin Pan and Yu Liu},
      year={2025},
      eprint={2503.07598},
      archivePrefix={arXiv},
      primaryClass={cs.CV},
      url={https://arxiv.org/abs/2503.07598}, 
}

@misc{wan,
      title={Wan: Open and Advanced Large-Scale Video Generative Models}, 
      author={Team Wan and Ang Wang and Baole Ai and Bin Wen and Chaojie Mao and Chen-Wei Xie and Di Chen and Feiwu Yu and Haiming Zhao and Jianxiao Yang and Jianyuan Zeng and Jiayu Wang and Jingfeng Zhang and Jingren Zhou and Jinkai Wang and Jixuan Chen and Kai Zhu and Kang Zhao and Keyu Yan and Lianghua Huang and Mengyang Feng and Ningyi Zhang and Pandeng Li and Pingyu Wu and Ruihang Chu and Ruili Feng and Shiwei Zhang and Siyang Sun and Tao Fang and Tianxing Wang and Tianyi Gui and Tingyu Weng and Tong Shen and Wei Lin and Wei Wang and Wei Wang and Wenmeng Zhou and Wente Wang and Wenting Shen and Wenyuan Yu and Xianzhong Shi and Xiaoming Huang and Xin Xu and Yan Kou and Yangyu Lv and Yifei Li and Yijing Liu and Yiming Wang and Yingya Zhang and Yitong Huang and Yong Li and You Wu and Yu Liu and Yulin Pan and Yun Zheng and Yuntao Hong and Yupeng Shi and Yutong Feng and Zeyinzi Jiang and Zhen Han and Zhi-Fan Wu and Ziyu Liu},
      year={2025},
      eprint={2503.20314},
      archivePrefix={arXiv},
      primaryClass={cs.CV},
      url={https://arxiv.org/abs/2503.20314}, 
}

@misc{rt1,
      title={RT-1: Robotics Transformer for Real-World Control at Scale}, 
      author={Anthony Brohan and Noah Brown and Justice Carbajal and Yevgen Chebotar and Joseph Dabis and Chelsea Finn and Keerthana Gopalakrishnan and Karol Hausman and Alex Herzog and Jasmine Hsu and Julian Ibarz and Brian Ichter and Alex Irpan and Tomas Jackson and Sally Jesmonth and Nikhil J Joshi and Ryan Julian and Dmitry Kalashnikov and Yuheng Kuang and Isabel Leal and Kuang-Huei Lee and Sergey Levine and Yao Lu and Utsav Malla and Deeksha Manjunath and Igor Mordatch and Ofir Nachum and Carolina Parada and Jodilyn Peralta and Emily Perez and Karl Pertsch and Jornell Quiambao and Kanishka Rao and Michael Ryoo and Grecia Salazar and Pannag Sanketi and Kevin Sayed and Jaspiar Singh and Sumedh Sontakke and Austin Stone and Clayton Tan and Huong Tran and Vincent Vanhoucke and Steve Vega and Quan Vuong and Fei Xia and Ted Xiao and Peng Xu and Sichun Xu and Tianhe Yu and Brianna Zitkovich},
      year={2023},
      eprint={2212.06817},
      archivePrefix={arXiv},
      primaryClass={cs.RO},
      url={https://arxiv.org/abs/2212.06817}, 
}

@misc{openxembodiment,
      title={Open X-Embodiment: Robotic Learning Datasets and RT-X Models}, 
      author={Embodiment Collaboration and Abby O'Neill and Abdul Rehman and Abhinav Gupta and Abhiram Maddukuri and Abhishek Gupta and Abhishek Padalkar and Abraham Lee and Acorn Pooley and Agrim Gupta and Ajay Mandlekar and Ajinkya Jain and Albert Tung and Alex Bewley and Alex Herzog and Alex Irpan and Alexander Khazatsky and Anant Rai and Anchit Gupta and Andrew Wang and Andrey Kolobov and Anikait Singh and Animesh Garg and Aniruddha Kembhavi and Annie Xie and Anthony Brohan and Antonin Raffin and Archit Sharma and Arefeh Yavary and Arhan Jain and Ashwin Balakrishna and Ayzaan Wahid and Ben Burgess-Limerick and Beomjoon Kim and Bernhard Schölkopf and Blake Wulfe and Brian Ichter and Cewu Lu and Charles Xu and Charlotte Le and Chelsea Finn and Chen Wang and Chenfeng Xu and Cheng Chi and Chenguang Huang and Christine Chan and Christopher Agia and Chuer Pan and Chuyuan Fu and Coline Devin and Danfei Xu and Daniel Morton and Danny Driess and Daphne Chen and Deepak Pathak and Dhruv Shah and Dieter Büchler and Dinesh Jayaraman and Dmitry Kalashnikov and Dorsa Sadigh and Edward Johns and Ethan Foster and Fangchen Liu and Federico Ceola and Fei Xia and Feiyu Zhao and Felipe Vieira Frujeri and Freek Stulp and Gaoyue Zhou and Gaurav S. Sukhatme and Gautam Salhotra and Ge Yan and Gilbert Feng and Giulio Schiavi and Glen Berseth and Gregory Kahn and Guangwen Yang and Guanzhi Wang and Hao Su and Hao-Shu Fang and Haochen Shi and Henghui Bao and Heni Ben Amor and Henrik I Christensen and Hiroki Furuta and Homanga Bharadhwaj and Homer Walke and Hongjie Fang and Huy Ha and Igor Mordatch and Ilija Radosavovic and Isabel Leal and Jacky Liang and Jad Abou-Chakra and Jaehyung Kim and Jaimyn Drake and Jan Peters and Jan Schneider and Jasmine Hsu and Jay Vakil and Jeannette Bohg and Jeffrey Bingham and Jeffrey Wu and Jensen Gao and Jiaheng Hu and Jiajun Wu and Jialin Wu and Jiankai Sun and Jianlan Luo and Jiayuan Gu and Jie Tan and Jihoon Oh and Jimmy Wu and Jingpei Lu and Jingyun Yang and Jitendra Malik and João Silvério and Joey Hejna and Jonathan Booher and Jonathan Tompson and Jonathan Yang and Jordi Salvador and Joseph J. Lim and Junhyek Han and Kaiyuan Wang and Kanishka Rao and Karl Pertsch and Karol Hausman and Keegan Go and Keerthana Gopalakrishnan and Ken Goldberg and Kendra Byrne and Kenneth Oslund and Kento Kawaharazuka and Kevin Black and Kevin Lin and Kevin Zhang and Kiana Ehsani and Kiran Lekkala and Kirsty Ellis and Krishan Rana and Krishnan Srinivasan and Kuan Fang and Kunal Pratap Singh and Kuo-Hao Zeng and Kyle Hatch and Kyle Hsu and Laurent Itti and Lawrence Yunliang Chen and Lerrel Pinto and Li Fei-Fei and Liam Tan and Linxi "Jim" Fan and Lionel Ott and Lisa Lee and Luca Weihs and Magnum Chen and Marion Lepert and Marius Memmel and Masayoshi Tomizuka and Masha Itkina and Mateo Guaman Castro and Max Spero and Maximilian Du and Michael Ahn and Michael C. Yip and Mingtong Zhang and Mingyu Ding and Minho Heo and Mohan Kumar Srirama and Mohit Sharma and Moo Jin Kim and Muhammad Zubair Irshad and Naoaki Kanazawa and Nicklas Hansen and Nicolas Heess and Nikhil J Joshi and Niko Suenderhauf and Ning Liu and Norman Di Palo and Nur Muhammad Mahi Shafiullah and Oier Mees and Oliver Kroemer and Osbert Bastani and Pannag R Sanketi and Patrick "Tree" Miller and Patrick Yin and Paul Wohlhart and Peng Xu and Peter David Fagan and Peter Mitrano and Pierre Sermanet and Pieter Abbeel and Priya Sundaresan and Qiuyu Chen and Quan Vuong and Rafael Rafailov and Ran Tian and Ria Doshi and Roberto Martín-Martín and Rohan Baijal and Rosario Scalise and Rose Hendrix and Roy Lin and Runjia Qian and Ruohan Zhang and Russell Mendonca and Rutav Shah and Ryan Hoque and Ryan Julian and Samuel Bustamante and Sean Kirmani and Sergey Levine and Shan Lin and Sherry Moore and Shikhar Bahl and Shivin Dass and Shubham Sonawani and Shubham Tulsiani and Shuran Song and Sichun Xu and Siddhant Haldar and Siddharth Karamcheti and Simeon Adebola and Simon Guist and Soroush Nasiriany and Stefan Schaal and Stefan Welker and Stephen Tian and Subramanian Ramamoorthy and Sudeep Dasari and Suneel Belkhale and Sungjae Park and Suraj Nair and Suvir Mirchandani and Takayuki Osa and Tanmay Gupta and Tatsuya Harada and Tatsuya Matsushima and Ted Xiao and Thomas Kollar and Tianhe Yu and Tianli Ding and Todor Davchev and Tony Z. Zhao and Travis Armstrong and Trevor Darrell and Trinity Chung and Vidhi Jain and Vikash Kumar and Vincent Vanhoucke and Vitor Guizilini and Wei Zhan and Wenxuan Zhou and Wolfram Burgard and Xi Chen and Xiangyu Chen and Xiaolong Wang and Xinghao Zhu and Xinyang Geng and Xiyuan Liu and Xu Liangwei and Xuanlin Li and Yansong Pang and Yao Lu and Yecheng Jason Ma and Yejin Kim and Yevgen Chebotar and Yifan Zhou and Yifeng Zhu and Yilin Wu and Ying Xu and Yixuan Wang and Yonatan Bisk and Yongqiang Dou and Yoonyoung Cho and Youngwoon Lee and Yuchen Cui and Yue Cao and Yueh-Hua Wu and Yujin Tang and Yuke Zhu and Yunchu Zhang and Yunfan Jiang and Yunshuang Li and Yunzhu Li and Yusuke Iwasawa and Yutaka Matsuo and Zehan Ma and Zhuo Xu and Zichen Jeff Cui and Zichen Zhang and Zipeng Fu and Zipeng Lin},
      year={2025},
      eprint={2310.08864},
      archivePrefix={arXiv},
      primaryClass={cs.RO},
      url={https://arxiv.org/abs/2310.08864}, 
}

@misc{cosmospredict25,
      title={World Simulation with Video Foundation Models for Physical AI}, 
      author={NVIDIA and : and Arslan Ali and Junjie Bai and Maciej Bala and Yogesh Balaji and Aaron Blakeman and Tiffany Cai and Jiaxin Cao and Tianshi Cao and Elizabeth Cha and Yu-Wei Chao and Prithvijit Chattopadhyay and Mike Chen and Yongxin Chen and Yu Chen and Shuai Cheng and Yin Cui and Jenna Diamond and Yifan Ding and Jiaojiao Fan and Linxi Fan and Liang Feng and Francesco Ferroni and Sanja Fidler and Xiao Fu and Ruiyuan Gao and Yunhao Ge and Jinwei Gu and Aryaman Gupta and Siddharth Gururani and Imad El Hanafi and Ali Hassani and Zekun Hao and Jacob Huffman and Joel Jang and Pooya Jannaty and Jan Kautz and Grace Lam and Xuan Li and Zhaoshuo Li and Maosheng Liao and Chen-Hsuan Lin and Tsung-Yi Lin and Yen-Chen Lin and Huan Ling and Ming-Yu Liu and Xian Liu and Yifan Lu and Alice Luo and Qianli Ma and Hanzi Mao and Kaichun Mo and Seungjun Nah and Yashraj Narang and Abhijeet Panaskar and Lindsey Pavao and Trung Pham and Morteza Ramezanali and Fitsum Reda and Scott Reed and Xuanchi Ren and Haonan Shao and Yue Shen and Stella Shi and Shuran Song and Bartosz Stefaniak and Shangkun Sun and Shitao Tang and Sameena Tasmeen and Lyne Tchapmi and Wei-Cheng Tseng and Jibin Varghese and Andrew Z. Wang and Hao Wang and Haoxiang Wang and Heng Wang and Ting-Chun Wang and Fangyin Wei and Jiashu Xu and Dinghao Yang and Xiaodong Yang and Haotian Ye and Seonghyeon Ye and Xiaohui Zeng and Jing Zhang and Qinsheng Zhang and Kaiwen Zheng and Andrew Zhu and Yuke Zhu},
      year={2025},
      eprint={2511.00062},
      archivePrefix={arXiv},
      primaryClass={cs.CV},
      url={https://arxiv.org/abs/2511.00062}, 
}

@misc{dreamtocontrol,
      title={Dream to Control: Learning Behaviors by Latent Imagination}, 
      author={Danijar Hafner and Timothy Lillicrap and Jimmy Ba and Mohammad Norouzi},
      year={2020},
      eprint={1912.01603},
      archivePrefix={arXiv},
      primaryClass={cs.LG},
      url={https://arxiv.org/abs/1912.01603}, 
}

@misc{cosmos,
      title={Cosmos World Foundation Model Platform for Physical AI}, 
      author={NVIDIA and : and Niket Agarwal and Arslan Ali and Maciej Bala and Yogesh Balaji and Erik Barker and Tiffany Cai and Prithvijit Chattopadhyay and Yongxin Chen and Yin Cui and Yifan Ding and Daniel Dworakowski and Jiaojiao Fan and Michele Fenzi and Francesco Ferroni and Sanja Fidler and Dieter Fox and Songwei Ge and Yunhao Ge and Jinwei Gu and Siddharth Gururani and Ethan He and Jiahui Huang and Jacob Huffman and Pooya Jannaty and Jingyi Jin and Seung Wook Kim and Gergely Klár and Grace Lam and Shiyi Lan and Laura Leal-Taixe and Anqi Li and Zhaoshuo Li and Chen-Hsuan Lin and Tsung-Yi Lin and Huan Ling and Ming-Yu Liu and Xian Liu and Alice Luo and Qianli Ma and Hanzi Mao and Kaichun Mo and Arsalan Mousavian and Seungjun Nah and Sriharsha Niverty and David Page and Despoina Paschalidou and Zeeshan Patel and Lindsey Pavao and Morteza Ramezanali and Fitsum Reda and Xiaowei Ren and Vasanth Rao Naik Sabavat and Ed Schmerling and Stella Shi and Bartosz Stefaniak and Shitao Tang and Lyne Tchapmi and Przemek Tredak and Wei-Cheng Tseng and Jibin Varghese and Hao Wang and Haoxiang Wang and Heng Wang and Ting-Chun Wang and Fangyin Wei and Xinyue Wei and Jay Zhangjie Wu and Jiashu Xu and Wei Yang and Lin Yen-Chen and Xiaohui Zeng and Yu Zeng and Jing Zhang and Qinsheng Zhang and Yuxuan Zhang and Qingqing Zhao and Artur Zolkowski},
      year={2025},
      eprint={2501.03575},
      archivePrefix={arXiv},
      primaryClass={cs.CV},
      url={https://arxiv.org/abs/2501.03575}, 
}

@misc{kling,
  author       = {Kuaishou},
  title        = {Kling},
  year         = {2024},
  howpublished = {\url{https://klingai.com/}},
  note         = {Accessed: 2025-11-08}
}

@misc{aleph,
  author       = {Runway},
  title        = {Runway Aleph},
  year         = {2025},
  howpublished = {\url{https://runwayml.com/research/introducing-runway-aleph}},
  note         = {Accessed: 2025-11-08}
}

@misc{dreamgen,
      title={DreamGen: Unlocking Generalization in Robot Learning through Video World Models}, 
      author={Joel Jang and Seonghyeon Ye and Zongyu Lin and Jiannan Xiang and Johan Bjorck and Yu Fang and Fengyuan Hu and Spencer Huang and Kaushil Kundalia and Yen-Chen Lin and Loic Magne and Ajay Mandlekar and Avnish Narayan and You Liang Tan and Guanzhi Wang and Jing Wang and Qi Wang and Yinzhen Xu and Xiaohui Zeng and Kaiyuan Zheng and Ruijie Zheng and Ming-Yu Liu and Luke Zettlemoyer and Dieter Fox and Jan Kautz and Scott Reed and Yuke Zhu and Linxi Fan},
      year={2025},
      eprint={2505.12705},
      archivePrefix={arXiv},
      primaryClass={cs.RO},
      url={https://arxiv.org/abs/2505.12705}, 
}

@misc{ego4d,
      title={Ego4D: Around the World in 3,000 Hours of Egocentric Video}, 
      author={Kristen Grauman and Andrew Westbury and Eugene Byrne and Zachary Chavis and Antonino Furnari and Rohit Girdhar and Jackson Hamburger and Hao Jiang and Miao Liu and Xingyu Liu and Miguel Martin and Tushar Nagarajan and Ilija Radosavovic and Santhosh Kumar Ramakrishnan and Fiona Ryan and Jayant Sharma and Michael Wray and Mengmeng Xu and Eric Zhongcong Xu and Chen Zhao and Siddhant Bansal and Dhruv Batra and Vincent Cartillier and Sean Crane and Tien Do and Morrie Doulaty and Akshay Erapalli and Christoph Feichtenhofer and Adriano Fragomeni and Qichen Fu and Abrham Gebreselasie and Cristina Gonzalez and James Hillis and Xuhua Huang and Yifei Huang and Wenqi Jia and Weslie Khoo and Jachym Kolar and Satwik Kottur and Anurag Kumar and Federico Landini and Chao Li and Yanghao Li and Zhenqiang Li and Karttikeya Mangalam and Raghava Modhugu and Jonathan Munro and Tullie Murrell and Takumi Nishiyasu and Will Price and Paola Ruiz Puentes and Merey Ramazanova and Leda Sari and Kiran Somasundaram and Audrey Southerland and Yusuke Sugano and Ruijie Tao and Minh Vo and Yuchen Wang and Xindi Wu and Takuma Yagi and Ziwei Zhao and Yunyi Zhu and Pablo Arbelaez and David Crandall and Dima Damen and Giovanni Maria Farinella and Christian Fuegen and Bernard Ghanem and Vamsi Krishna Ithapu and C. V. Jawahar and Hanbyul Joo and Kris Kitani and Haizhou Li and Richard Newcombe and Aude Oliva and Hyun Soo Park and James M. Rehg and Yoichi Sato and Jianbo Shi and Mike Zheng Shou and Antonio Torralba and Lorenzo Torresani and Mingfei Yan and Jitendra Malik},
      year={2022},
      eprint={2110.07058},
      archivePrefix={arXiv},
      primaryClass={cs.CV},
      url={https://arxiv.org/abs/2110.07058}, 
}

@misc{masquerade,
      title={Masquerade: Learning from In-the-wild Human Videos using Data-Editing}, 
      author={Marion Lepert and Jiaying Fang and Jeannette Bohg},
      year={2025},
      eprint={2508.09976},
      archivePrefix={arXiv},
      primaryClass={cs.RO},
      url={https://arxiv.org/abs/2508.09976}, 
}

@article{hrp,
  title={Hrp: Human affordances for robotic pre-training},
  author={Srirama, Mohan Kumar and Dasari, Sudeep and Bahl, Shikhar and Gupta, Abhinav},
  journal={arXiv preprint arXiv:2407.18911},
  year={2024}
}

@misc{human2robot,
      title={Human2Robot: Learning Robot Actions from Paired Human-Robot Videos}, 
      author={Sicheng Xie and Haidong Cao and Zejia Weng and Zhen Xing and Haoran Chen and Shiwei Shen and Jiaqi Leng and Zuxuan Wu and Yu-Gang Jiang},
      year={2025},
      eprint={2502.16587},
      archivePrefix={arXiv},
      primaryClass={cs.RO},
      url={https://arxiv.org/abs/2502.16587}, 
}

@misc{egomimic,
      title={EgoMimic: Scaling Imitation Learning via Egocentric Video}, 
      author={Simar Kareer and Dhruv Patel and Ryan Punamiya and Pranay Mathur and Shuo Cheng and Chen Wang and Judy Hoffman and Danfei Xu},
      year={2024},
      eprint={2410.24221},
      archivePrefix={arXiv},
      primaryClass={cs.RO},
      url={https://arxiv.org/abs/2410.24221}, 
}

@misc{egozero,
      title={EgoZero: Robot Learning from Smart Glasses}, 
      author={Vincent Liu and Ademi Adeniji and Haotian Zhan and Siddhant Haldar and Raunaq Bhirangi and Pieter Abbeel and Lerrel Pinto},
      year={2025},
      eprint={2505.20290},
      archivePrefix={arXiv},
      primaryClass={cs.RO},
      url={https://arxiv.org/abs/2505.20290}, 
}

@inproceedings{videodex,
  title={Videodex: Learning dexterity from internet videos},
  author={Shaw, Kenneth and Bahl, Shikhar and Pathak, Deepak},
  booktitle={Conference on Robot Learning},
  pages={654--665},
  year={2023},
  organization={PMLR}
}

@inproceedings{dexvip,
  title={Dexvip: Learning dexterous grasping with human hand pose priors from video},
  author={Mandikal, Priyanka and Grauman, Kristen},
  booktitle={Conference on Robot Learning},
  pages={651--661},
  year={2022},
  organization={PMLR}
}

@inproceedings{track2act,
  title={Track2act: Predicting point tracks from internet videos enables generalizable robot manipulation},
  author={Bharadhwaj, Homanga and Mottaghi, Roozbeh and Gupta, Abhinav and Tulsiani, Shubham},
  booktitle={European Conference on Computer Vision},
  pages={306--324},
  year={2024},
  organization={Springer}
}

@misc{h2r,
      title={H2R: A Human-to-Robot Data Augmentation for Robot Pre-training from Videos}, 
      author={Guangrun Li and Yaoxu Lyu and Zhuoyang Liu and Chengkai Hou and Jieyu Zhang and Shanghang Zhang},
      year={2025},
      eprint={2505.11920},
      archivePrefix={arXiv},
      primaryClass={cs.RO},
      url={https://arxiv.org/abs/2505.11920}, 
}

@misc{immimic,
      title={ImMimic: Cross-Domain Imitation from Human Videos via Mapping and Interpolation}, 
      author={Yangcen Liu and Woo Chul Shin and Yunhai Han and Zhenyang Chen and Harish Ravichandar and Danfei Xu},
      year={2025},
      eprint={2509.10952},
      archivePrefix={arXiv},
      primaryClass={cs.RO},
      url={https://arxiv.org/abs/2509.10952}, 
}

@misc{phantom,
      title={Phantom: Training Robots Without Robots Using Only Human Videos}, 
      author={Marion Lepert and Jiaying Fang and Jeannette Bohg},
      year={2025},
      eprint={2503.00779},
      archivePrefix={arXiv},
      primaryClass={cs.RO},
      url={https://arxiv.org/abs/2503.00779}, 
}

@article{droid,
  title={Droid: A large-scale in-the-wild robot manipulation dataset},
  author={Khazatsky, Alexander and Pertsch, Karl and Nair, Suraj and Balakrishna, Ashwin and Dasari, Sudeep and Karamcheti, Siddharth and Nasiriany, Soroush and Srirama, Mohan Kumar and Chen, Lawrence Yunliang and Ellis, Kirsty and others},
  journal={arXiv preprint arXiv:2403.12945},
  year={2024}
}

@inproceedings{bridgev2,
    title={BridgeData V2: A Dataset for Robot Learning at Scale},
    author={Walke, Homer and Black, Kevin and Lee, Abraham and Kim, Moo Jin and Du, Max and Zheng, Chongyi and Zhao, Tony and Hansen-Estruch, Philippe and Vuong, Quan and He, Andre and Myers, Vivek and Fang, Kuan and Finn, Chelsea and Levine, Sergey},
    booktitle={Conference on Robot Learning (CoRL)},
    year={2023}
}

@article{minimaxremover,
  title={MiniMax-Remover: Taming Bad Noise Helps Video Object Removal},
  author={Zi, Bojia and Peng, Weixuan and Qi, Xianbiao and Wang, Jianan and Zhao, Shihao and Xiao, Rong and Wong, Kam-Fai},
  journal={arXiv preprint arXiv:2505.24873},
  year={2025}
}

@inproceedings{e2fgvi,
  title={Towards an end-to-end framework for flow-guided video inpainting},
  author={Li, Zhen and Lu, Cheng-Ze and Qin, Jianhua and Guo, Chun-Le and Cheng, Ming-Ming},
  booktitle={Proceedings of the IEEE/CVF conference on computer vision and pattern recognition},
  pages={17562--17571},
  year={2022}
}

@misc{gsam,
      title={Grounded SAM: Assembling Open-World Models for Diverse Visual Tasks}, 
      author={Tianhe Ren and Shilong Liu and Ailing Zeng and Jing Lin and Kunchang Li and He Cao and Jiayu Chen and Xinyu Huang and Yukang Chen and Feng Yan and Zhaoyang Zeng and Hao Zhang and Feng Li and Jie Yang and Hongyang Li and Qing Jiang and Lei Zhang},
      year={2024},
      eprint={2401.14159},
      archivePrefix={arXiv},
      primaryClass={cs.CV}
}

@misc{sam2,
      title={SAM 2: Segment Anything in Images and Videos}, 
      author={Nikhila Ravi and Valentin Gabeur and Yuan-Ting Hu and Ronghang Hu and Chaitanya Ryali and Tengyu Ma and Haitham Khedr and Roman Rädle and Chloe Rolland and Laura Gustafson and Eric Mintun and Junting Pan and Kalyan Vasudev Alwala and Nicolas Carion and Chao-Yuan Wu and Ross Girshick and Piotr Dollár and Christoph Feichtenhofer},
      year={2024},
      eprint={2408.00714},
      archivePrefix={arXiv},
      primaryClass={cs.CV},
      url={https://arxiv.org/abs/2408.00714}, 
}

@article{lora,
  title={Lora: Low-rank adaptation of large language models.},
  author={Hu, Edward J and Shen, Yelong and Wallis, Phillip and Allen-Zhu, Zeyuan and Li, Yuanzhi and Wang, Shean and Wang, Lu and Chen, Weizhu and others},
  journal={ICLR},
  volume={1},
  number={2},
  pages={3},
  year={2022}
}

@inproceedings{hamer,
    title={Reconstructing Hands in 3{D} with Transformers},
    author={Pavlakos, Georgios and Shan, Dandan and Radosavovic, Ilija and Kanazawa, Angjoo and Fouhey, David and Malik, Jitendra},
    booktitle={CVPR},
    year={2024}
}

@inproceedings{
  vitpose,
  title={Vi{TP}ose: Simple Vision Transformer Baselines for Human Pose Estimation},
  author={Yufei Xu and Jing Zhang and Qiming Zhang and Dacheng Tao},
  booktitle={Advances in Neural Information Processing Systems},
  year={2022},
}

@article{pi05,
  title={pi0.5: a Vision-Language-Action Model with Open-World Generalization},
  author={Intelligence, Physical and Black, Kevin and Brown, Noah and Darpinian, James and Dhabalia, Karan and Driess, Danny and Esmail, Adnan and Equi, Michael and Finn, Chelsea and Fusai, Niccolo and others},
  journal={arXiv preprint arXiv:2504.16054},
  year={2025}
}

@inproceedings{hopman,
  title={Towards generalizable zero-shot manipulation via translating human interaction plans},
  author={Bharadhwaj, Homanga and Gupta, Abhinav and Kumar, Vikash and Tulsiani, Shubham},
  booktitle={2024 IEEE International Conference on Robotics and Automation (ICRA)},
  pages={6904--6911},
  year={2024},
  organization={IEEE}
}

@article{hrdt,
  title={H-rdt: Human manipulation enhanced bimanual robotic manipulation},
  author={Bi, Hongzhe and Wu, Lingxuan and Lin, Tianwei and Tan, Hengkai and Su, Zhizhong and Su, Hang and Zhu, Jun},
  journal={arXiv preprint arXiv:2507.23523},
  year={2025}
}

@article{hat,
  title={Humanoid Policy\~{} Human Policy},
  author={Qiu, Ri-Zhao and Yang, Shiqi and Cheng, Xuxin and Chawla, Chaitanya and Li, Jialong and He, Tairan and Yan, Ge and Yoon, David J and Hoque, Ryan and Paulsen, Lars and others},
  journal={arXiv preprint arXiv:2503.13441},
  year={2025}
}

@article{manipulatorindep,
  title={Manipulator-independent representations for visual imitation},
  author={Zhou, Yuxiang and Aytar, Yusuf and Bousmalis, Konstantinos},
  journal={arXiv preprint arXiv:2103.09016},
  year={2021}
}

@article{mendonca2023structured,
  title={Structured world models from human videos},
  author={Mendonca, Russell and Bahl, Shikhar and Pathak, Deepak},
  journal={arXiv preprint arXiv:2308.10901},
  year={2023}
}

@inproceedings{bahl2023affordances,
  title={Affordances from human videos as a versatile representation for robotics},
  author={Bahl, Shikhar and Mendonca, Russell and Chen, Lili and Jain, Unnat and Pathak, Deepak},
  booktitle={Proceedings of the IEEE/CVF Conference on Computer Vision and Pattern Recognition},
  pages={13778--13790},
  year={2023}
}

@article{bahl2022human,
  title={Human-to-robot imitation in the wild},
  author={Bahl, Shikhar and Gupta, Abhinav and Pathak, Deepak},
  journal={arXiv preprint arXiv:2207.09450},
  year={2022}
}

@article{wen2023any,
  title={Any-point trajectory modeling for policy learning},
  author={Wen, Chuan and Lin, Xingyu and So, John and Chen, Kai and Dou, Qi and Gao, Yang and Abbeel, Pieter},
  journal={arXiv preprint arXiv:2401.00025},
  year={2023}
}

@article{li2024okami,
  title={Okami: Teaching humanoid robots manipulation skills through single video imitation},
  author={Li, Jinhan and Zhu, Yifeng and Xie, Yuqi and Jiang, Zhenyu and Seo, Mingyo and Pavlakos, Georgios and Zhu, Yuke},
  journal={arXiv preprint arXiv:2410.11792},
  year={2024}
}

@inproceedings{das2021model,
  title={Model-based inverse reinforcement learning from visual demonstrations},
  author={Das, Neha and Bechtle, Sarah and Davchev, Todor and Jayaraman, Dinesh and Rai, Akshara and Meier, Franziska},
  booktitle={Conference on Robot Learning},
  pages={1930--1942},
  year={2021},
  organization={PMLR}
}

@inproceedings{xiong2021learning,
  title={Learning by watching: Physical imitation of manipulation skills from human videos},
  author={Xiong, Haoyu and Li, Quanzhou and Chen, Yun-Chun and Bharadhwaj, Homanga and Sinha, Samarth and Garg, Animesh},
  booktitle={2021 IEEE/RSJ international conference on intelligent robots and systems (iros)},
  pages={7827--7834},
  year={2021},
  organization={IEEE}
}

@inproceedings{goyal2022ifor,
  title={Ifor: Iterative flow minimization for robotic object rearrangement},
  author={Goyal, Ankit and Mousavian, Arsalan and Paxton, Chris and Chao, Yu-Wei and Okorn, Brian and Deng, Jia and Fox, Dieter},
  booktitle={Proceedings of the IEEE/CVF conference on computer vision and pattern recognition},
  pages={14787--14797},
  year={2022}
}

@inproceedings{seita2023toolflownet,
  title={Toolflownet: Robotic manipulation with tools via predicting tool flow from point clouds},
  author={Seita, Daniel and Wang, Yufei and Shetty, Sarthak J and Li, Edward Yao and Erickson, Zackory and Held, David},
  booktitle={Conference on Robot Learning},
  pages={1038--1049},
  year={2023},
  organization={PMLR}
}

@article{shadow,
  title={Shadow: Leveraging segmentation masks for cross-embodiment policy transfer},
  author={Lepert, Marion and Doshi, Ria and Bohg, Jeannette},
  journal={arXiv preprint arXiv:2503.00774},
  year={2025}
}

@article{roviaug,
  title={Rovi-aug: Robot and viewpoint augmentation for cross-embodiment robot learning},
  author={Chen, Lawrence Yunliang and Xu, Chenfeng and Dharmarajan, Karthik and Irshad, Muhammad Zubair and Cheng, Richard and Keutzer, Kurt and Tomizuka, Masayoshi and Vuong, Quan and Goldberg, Ken},
  journal={arXiv preprint arXiv:2409.03403},
  year={2024}
}

@ARTICLE{epickitchen,
           title={Rescaling Egocentric Vision: Collection, Pipeline and Challenges for EPIC-KITCHENS-100},
           author={Damen, Dima and Doughty, Hazel and Farinella, Giovanni Maria and Furnari, Antonino 
           and Ma, Jian and Kazakos, Evangelos and Moltisanti, Davide and Munro, Jonathan 
           and Perrett, Toby and Price, Will and Wray, Michael},
           journal   = {International Journal of Computer Vision (IJCV)},
           year      = {2022},
           volume = {130},
           pages = {33–55},
           Url       = {https://doi.org/10.1007/s11263-021-01531-2}
}

@inproceedings{affordancediffusion,
  title={Affordance diffusion: Synthesizing hand-object interactions},
  author={Ye, Yufei and Li, Xueting and Gupta, Abhinav and De Mello, Shalini and Birchfield, Stan and Song, Jiaming and Tulsiani, Shubham and Liu, Sifei},
  booktitle={Proceedings of the IEEE/CVF Conference on Computer Vision and Pattern Recognition},
  pages={22479--22489},
  year={2023}
}

@article{robomaster,
  title={Learning Video Generation for Robotic Manipulation with Collaborative Trajectory Control},
  author={Fu, Xiao and Wang, Xintao and Liu, Xian and Bai, Jianhong and Xu, Runsen and Wan, Pengfei and Zhang, Di and Lin, Dahua},
  journal={arXiv preprint arXiv:2506.01943},
  year={2025}
}

@article{irasim,
  title={Irasim: Learning interactive real-robot action simulators},
  author={Zhu, Fangqi and Wu, Hongtao and Guo, Song and Liu, Yuxiao and Cheang, Chilam and Kong, Tao},
  journal={arXiv preprint arXiv:2406.14540},
  year={2024}
}

@article{ivideogpt,
  title={ivideogpt: Interactive videogpts are scalable world models},
  author={Wu, Jialong and Yin, Shaofeng and Feng, Ningya and He, Xu and Li, Dong and Hao, Jianye and Long, Mingsheng},
  journal={Advances in Neural Information Processing Systems},
  volume={37},
  pages={68082--68119},
  year={2024}
}

@article{hma,
  title={Learning Real-World Action-Video Dynamics with Heterogeneous Masked Autoregression},
  author={Wang, Lirui and Zhao, Kevin and Liu, Chaoqi and Chen, Xinlei},
  journal={arXiv preprint arXiv:2502.04296},
  year={2025}
}

@article{uva,
  title={Unified video action model},
  author={Li, Shuang and Gao, Yihuai and Sadigh, Dorsa and Song, Shuran},
  journal={arXiv preprint arXiv:2503.00200},
  year={2025}
}

@article{worldvla,
  title={WorldVLA: Towards Autoregressive Action World Model},
  author={Cen, Jun and Yu, Chaohui and Yuan, Hangjie and Jiang, Yuming and Huang, Siteng and Guo, Jiayan and Li, Xin and Song, Yibing and Luo, Hao and Wang, Fan and others},
  journal={arXiv preprint arXiv:2506.21539},
  year={2025}
}

@article{unipi,
  title={Learning universal policies via text-guided video generation},
  author={Du, Yilun and Yang, Sherry and Dai, Bo and Dai, Hanjun and Nachum, Ofir and Tenenbaum, Josh and Schuurmans, Dale and Abbeel, Pieter},
  journal={Advances in neural information processing systems},
  volume={36},
  pages={9156--9172},
  year={2023}
}

@inproceedings{controlnet,
  title={Adding conditional control to text-to-image diffusion models},
  author={Zhang, Lvmin and Rao, Anyi and Agrawala, Maneesh},
  booktitle={Proceedings of the IEEE/CVF international conference on computer vision},
  pages={3836--3847},
  year={2023}
}

@article{qwen2,
  title={Qwen2. 5-vl technical report},
  author={Bai, Shuai and Chen, Keqin and Liu, Xuejing and Wang, Jialin and Ge, Wenbin and Song, Sibo and Dang, Kai and Wang, Peng and Wang, Shijie and Tang, Jun and others},
  journal={arXiv preprint arXiv:2502.13923},
  year={2025}
}

@inproceedings{dexycb,
  title={DexYCB: A benchmark for capturing hand grasping of objects},
  author={Chao, Yu-Wei and Yang, Wei and Xiang, Yu and Molchanov, Pavlo and Handa, Ankur and Tremblay, Jonathan and Narang, Yashraj S and Van Wyk, Karl and Iqbal, Umar and Birchfield, Stan and others},
  booktitle={Proceedings of the IEEE/CVF conference on computer vision and pattern recognition},
  pages={9044--9053},
  year={2021}
}

@inproceedings{lpips,
  title={The Unreasonable Effectiveness of Deep Features as a Perceptual Metric},
  author={Zhang, Richard and Isola, Phillip and Efros, Alexei A and Shechtman, Eli and Wang, Oliver},
  booktitle={CVPR},
  year={2018}
}

@inproceedings{cotracker3,
  title={Cotracker3: Simpler and better point tracking by pseudo-labelling real videos},
  author={Karaev, Nikita and Makarov, Yuri and Wang, Jianyuan and Neverova, Natalia and Vedaldi, Andrea and Rupprecht, Christian},
  booktitle={Proceedings of the IEEE/CVF International Conference on Computer Vision},
  pages={6013--6022},
  year={2025}
}

@misc{Franka,
  author       = {Franka Robotics GmbH},
  title        = {Franka Robotics — Homepage},
  howpublished = {\url{https://franka.de/}},
  year         = {2025},
  note         = {Accessed: 2025-11-14}
}

@article{gemini,
  title={Gemini 2.5: Pushing the frontier with advanced reasoning, multimodality, long context, and next generation agentic capabilities},
  author={Comanici, Gheorghe and Bieber, Eric and Schaekermann, Mike and Pasupat, Ice and Sachdeva, Noveen and Dhillon, Inderjit and Blistein, Marcel and Ram, Ori and Zhang, Dan and Rosen, Evan and others},
  journal={arXiv preprint arXiv:2507.06261},
  year={2025}
}

@article{pointpolicy,
  title={Point policy: Unifying observations and actions with key points for robot manipulation},
  author={Haldar, Siddhant and Pinto, Lerrel},
  journal={arXiv preprint arXiv:2502.20391},
  year={2025}
}

@article{li2025mimicdreamer,
  title={Mimicdreamer: Aligning human and robot demonstrations for scalable vla training},
  author={Li, Haoyun and Zhang, Ivan and Ouyang, Runqi and Wang, Xiaofeng and Zhu, Zheng and Yang, Zhiqin and Zhang, Zhentao and Wang, Boyuan and Ni, Chaojun and Qin, Wenkang and others},
  journal={arXiv preprint arXiv:2509.22199},
  year={2025}
}

@article{team2025gigaworld,
  title={GigaWorld-0: World Models as Data Engine to Empower Embodied AI},
  author={Team, GigaWorld and Ye, Angen and Wang, Boyuan and Ni, Chaojun and Huang, Guan and Zhao, Guosheng and Li, Haoyun and Zhu, Jiagang and Li, Kerui and Xu, Mengyuan and others},
  journal={arXiv preprint arXiv:2511.19861},
  year={2025}
}
}

% WARNING: do not forget to delete the supplementary pages from your submission 
\clearpage
\maketitlesupplementary

\section{Motivation of H2Rep}
\label{sec:supp_motivation_h2rep}
In this paper, our \textit{H2Rep} representation overlays the abstract pose sequence onto the background video using an $\alpha$-blending scheme. Another natural design is to treat pose and background as two separate video streams—one containing only the background, and the other containing only the pose rendered on a white or black canvas. This alternative preserves more disentangled information.

However, under an in-context generation framework, using dual video streams would effectively \emph{double} the input tokens, causing both computation and memory to scale quadratically (i.e., $4\times$). To balance efficiency and expressiveness, we adopt the $\alpha$-blended formulation: the pose is overlaid with controlled transparency so as to minimally affect background content while substantially reducing computational and memory costs. Moreover, this representation remains pixel-aligned with both the human reference and the final generated robot video, which facilitates learning for the video generator.

\section{Inference Efficiency}
\label{sec:supp_inference_speed}
Our 5B in-context model runs at about 13 seconds per frame, taking about 648 seconds to generate a 49-frame 704$\times$1280 video on a single H200 GPU, with a peak memory consumption of 63\,GB.

% \section{Video Example Viewer}
% \label{sec:supp_video_examples}
% We provide multiple video examples in the supplementary material, which can be viewed using the accompanying HTML viewer. If the HTML viewer encounters any issues, the videos can also be played directly from their respective folders.

\end{document}